\documentclass[10pt,twocolumn,letterpaper]{article}
\usepackage{cvpr}
\usepackage{times}
\usepackage{epsfig}
\usepackage{graphicx}

\usepackage{amsmath}
\usepackage{amssymb}
\usepackage{xcolor,colortbl}
\usepackage{makecell}
\usepackage{rotating}
\usepackage{epsfig}
\usepackage{graphicx}
\usepackage{amsmath}
\usepackage{amssymb}
\usepackage{amsfonts}
\usepackage{empheq}
\usepackage{framed, color}
\usepackage{xcolor}
\usepackage{graphics}
\usepackage{graphicx}
\usepackage[]{graphicx} % more modern
\usepackage{epsfig} % less modern
\usepackage{subfigure}
\usepackage{algorithm}
\usepackage{algorithmic}
\usepackage{stmaryrd}
\usepackage[mathscr]{eucal}
\usepackage{lineno}
\usepackage{color}
\usepackage{filecontents}
\usepackage{subfigure}
\usepackage{multirow}
\usepackage{filecontents}
\usepackage{verbatim} % begin{comment} end{comment}
\usepackage{tabularx}
\usepackage{lineno}
\usepackage{setspace}
\usepackage{multirow}
\usepackage{textcomp,booktabs}
\usepackage{caption}
\usepackage{epstopdf}

\def\M{{\bf M}}

\def\0{{\bf 0}}
\def\1{{\bf 1}}

\definecolor{red}{rgb}{0.95,0.4,0.4}
\definecolor{blue}{rgb}{0.4,0.4,0.95}
\definecolor{darkblue}{rgb}{0,0,0.8}
\definecolor{darkred}{rgb}{0.8,0,0}
\definecolor{darkgreen}{rgb}{0,0.5,0}
\definecolor{grey}{rgb}{0.6,0.6,0.6}
\definecolor{darkgrey}{rgb}{0.2,0.2,0.2}
\definecolor{lightgrey}{rgb}{0.85,0.85,0.85}
\definecolor{col1}{RGB}{232, 161, 148}
\definecolor{col2}{RGB}{148, 187, 232}

\newcommand\shu[1]{\textcolor{orange}{\bf Shu: [[#1]]}}

\usepackage[breaklinks=true,bookmarks=false]{hyperref}

\cvprfinalcopy % *** Uncomment this line for the final submission

\def\cvprPaperID{****} % *** Enter the CVPR Paper ID here

\begin{document}

%%%%%%%%% TITLE
\title{\emph{Celeganser}:
Automated Analysis of Nematode
Morphology and Age
%\\ through Segmentation,  Body Regression and Age Estimation
}

\author{Linfeng Wang\textsuperscript{$\dagger$}, 
\ \  Shu Kong\textsuperscript{$\ast$}, 
\ \  Zachary Pincus\textsuperscript{$\ddagger$}, 
\ \  Charless Fowlkes\textsuperscript{$\dagger$} 
\\
{\small \{\tt linfenw2, fowlkes\}@ics.uci.edu \ \  shuk@andrew.cmu.edu \ \  	zpincus@wustl.edu }\\
\textsuperscript{$\dagger$}University of California, Irvine   \ \ \ 
\textsuperscript{$\ast$}Carnegie Mellon University   \ \ \ 
\textsuperscript{$\dagger$}Washington University in St. Louis \\ \\
%\institute{$\ast$Carnegie Mellon University   \ \ \ \ \   $\dag$Argo AI  \\
% {\small \{\tt shuk, deva\}@andrew.cmu.edu} \\ \ \\
{[\href{https://github.com/DerrickWanglf/Celeganser-Automated-Analysis-of-Nematode-Morphology-and-Age}{Project Page}]} \ \
{[\href{https://github.com/DerrickWanglf/Celeganser-Automated-Analysis-of-Nematode-Morphology-and-Age}{Github}]}  \ \
{[\href{https://docs.google.com/presentation/d/1lciS29gSs9eLH0w4hWQXu1al_SYfMjxj0o9erMjkwi8/edit?usp=sharing}{Slides}]} \ \
% {[\href{http://www.cs.cmu.edu/~shuk/slides/pruos_poster.pdf}{Poster}]} \ \
% }
}

%\author{Linfeng Wang\\
%Institution1\\\
%\and

\maketitle
%\thispagestyle{empty}

%%%%%%%%% ABSTRACT
\begin{abstract}
   The nematode Caenorhabditis elegans ({C. elegans})
   serves as an important model organism in a wide variety of biological studies.
   In this paper we introduce a pipeline for automated analysis of C. elegans
   imagery for the purpose of studying life-span, health-span and the underlying
   genetic determinants of aging. Our system detects and segments the worm, and predicts
   body coordinates at each pixel location inside the worm. These coordinates
   provide dense correspondence across individual animals to allow for meaningful
   comparative analysis. We show that a model pre-trained to perform body-coordinate regression
   extracts rich features that can be used to predict the age of individual worms
   with high accuracy. This lays the ground for future research in quantifying
   the relation between organs' physiologic and biochemical state, and individual
   life/health-span.
%   As C. elegans is too small to allow for convenient visual analysis,
%   it becomes immediately rewarding to build automated computer-added systems
%   for high-throughput research with C. elegans.
%   In this paper,
%   we introduce a prototyped, automated system that assists C. elegans analysis
%   within the research of  life/health-span.
%   We call the system ``{\bf Celeganser}''.
%   We leverage multiple computer vision techniques,
%   consisting of
%   worm segmentation, body coordinate regression and age regression.
%   While the system is largely standalone,
%   %and even seemingly unnecessary for age estimation (which is trivial to solve with the date tag),
%   we note Celeganser exploits organ locations for body coordinate regression and age regression.
   %maximize the research potential of C. elegans, it is necessary to automatically obtain the C. elegans' posture, texture and organ information and estimate the age under a high-precision microscope. We proposed a robust U-net based pipeline for performing automatic individual-level segmentation and pose regression of C. elegans in microscope images. Based on the extracted pose and texture feature, we also performed the age estimation for C. elegans and ran the experiments for age determining factors among background, shape and texture in C. elegans.

\end{abstract}

%%%%%%%%% BODY TEXT
\section{Introduction}
\emph{C. elegans} is one of the most important invertebrate model organisms in biological
research. Landmark studies with C. elegans span various disciplines and fields, including
large scale functional characterization of genes~\cite{brenner1974genetics},
complete tracing of cell lineage in developmental~\cite{sulston1977post}, and mapping
whole-animal nervous system connectomes~\cite{cook2019whole}.
C. elegans also offer an ideal model for understanding variability across individuals
that result in different health and life span~\cite{pincus2016autofluorescence, pincus2016ageing}.
Even C. elegans, with their famously invariant pattern of embryonic development, experience large differences in individual lifespans. Raised in identical lab environments, genetically identical C. elegans show as much relative variability around their two-week lifespans as humans do around theirs of eighty years~\cite{pincus2016autofluorescence}.
We hope to learn why this is, what more-robust individuals have that their frailer siblings lack, and what that means for human health and aging~\cite{pincus2016autofluorescence}.

In this paper we describe a prototype pipeline for performing high-throughput longitudinal
analysis of individual worms. This automated ``C. elegans analyzer'', which we call
\emph{Celeganser} consists of three computer vision modules: worm segmentation which segments
the worm from the plate background,worm body coordinate regression that regress pixels on the
worm body to a pre-defined coordinate system, and age regression that predicts the worm's age
based on the segmented worm body. The problem of predicting age is of biological interest in
its own right as the ``physiological age'' of an animal does not always match the chronological
age. We further hypothesize that if an automated system (e.g., a machine learning model) can
regress the body coordinate and age well, it must learn to extract image features that relate
to the internal structure of the worm the worm, e.g., organ placement and status (e.g., disorder)
that correlates with lifespan and healthspan.

%Moreover, the segmentation step helps filter out background factor that may negatively impact,
%or provide artificial factors for body and age regression.

In this paper, we describe the technical aspects of our prototype system and conduct a
thorough experimental study to justify the design choices. To summarize our contribution:
%Further analysis on the correlation of worm body and early anomalous status discovery will be a part
%of our future work, and thus beyond the scope of this paper.
%To conclude our contribution:
\begin{enumerate}
  \item We propose \emph{Celeganser}, an automated system assisting analysis of C. elegans for studying lifespan and healthspan and validate the components of the system including segmentation, body-coordinate regression and age regression.
  \item We demonstrate that models pre-trained for body-coordinate regression extract features
  which are useful in predicting age.
  \item We carry out experiments to analyze the extent to which the worm shape and size, internal appearance and background environment are predictive of age. This provides a foundation for future research in lifespan and healthspan.
\end{enumerate}
We start with related work in Section~\ref{sec:related_work},
and elaborate on details of our analysis modules in Section~\ref{sec:celeganser}.
In Section~\ref{sec:exp} we describe a thorough experimental validation,
and conclude in Section~\ref{sec:conclusion}.

%\emph{Celeganser}
%
%high-throughput analysis endowed by computer vision and machine learning,
%
%The
%latter helps accelerating biomedical research on lifespan, healthspan, hypothesis testing of genetic
%interactions with aging.
%
%The latter helps accelerating
%biomedical research on lifespan, healthspan, hypothesis testing of genetic interactions with
%aging.

\section{Related Work}
\label{sec:related_work}

\subsection{Automated analysis of C. elegans}
C. elegans is a microscopic nematode that is used in a wide variety of biological studies
and provides one of the best model systems to study lifespan
development~\cite{kenyon1993c, kimura1997daf, gonczy2005asymmetric,kenyon2010genetics}
as their genetics and development is well characterized and the small size of C. elegans
makes it possible to grow them in large numbers. Acquiring high-resolution survival data
has lead to fruitful discoveries~\cite{curtsinger1992demography, vaupel1998biodemographic, mair2003demography, johnson2001age, baeriswyl2010modulation}.

However,
this also brings a challenge as manual observation can be tedious and time-consuming.
To fully realize the potential for high-throughput studies with strong statistical power,
attention has turned to building automated systems that acquire and analyze the worms
through automatic microscopic scanning~\cite{stroustrup2013caenorhabditis, pincus2016ageing}.
For example, the {\em lifespan machine} is an automated system for imaging large populations
of worms using inexpensive hardware (flatbed scanners) to allow for quantitative investigations
into the statistical structure of aging as a population~\cite{stroustrup2013caenorhabditis}.
In this paper, we focus on analyzing individual worm at higher resolution (5-10x magnification)
in order to to precisely segment the worm from the scanned image, regress body coordinate and
estimate its age. Our system is standalone and re-trainable, and produces rich outputs that
are useful for a range of downstream studies of individual differences in lifespan and the
rate of aging.

We note that a recently published paper~\cite{lin2020using} carried out a related study, training Convolutional Neural Networks (CNN) for age estimation with worm images as input. Our approach
differs in that we image worms automatically (rather than removing individuals and anesthetizing
for imaging) allowing us to train and evaluate on a dataset which is 10x larger. We also report
age predictions that are somewhat more accurate and can naturally handle a wider variety of poses.

%However, such a practice may be misleading as the image background can reveal worm age through the accumulated food relic, dirt and eggs. In contrast, by rigorously analyze how age estimation performs with background only, worm silhouette, segmented worm and so on, we derive rich conclusions and meaningful suggestions to the community.\shu{check here}
%\ccf{Predicting age from images using a CNN was also explored by recent work of...
%[[cite Lin et al TCCB2020 and discuss relation?]]}

\subsection{Convolutional Neural Networks}

Convolutional Neural Networks (CNN) have become one of the most successful
models in machine learning and computer vision for various applications
(see e.g., ~\cite{esteva2017dermatologist, coudray2018classification, sahraeian2019deep}),
owing to their flexibility and state-of-the-art performance on many
vision tasks~\cite{lecun1998gradient, krizhevsky2012imagenet, he2016deep}
especially when leveraging large-scale data (like ImageNet~\cite{deng2009imagenet}).
CNN architectures designed for whole image classification typical involve
multiple layers of spatial pooling and produce a single output.
For the purpose a producing dense, per-pixel predictions and maintaining
high spatial resolution, we adopt the U-shaped architecture of ~\cite{ronneberger2015u}.
The U-shape architecture include an encoder which is essentially a traditional
CNN (e.g., ResNet~\cite{he2016deep}),
and a decoder which has skip connections to the encoder layers and upsampling layers.

In addition to outputting discrete per-pixel class labels (which we use for worm segmentation),
CNNs can be also trained to output continuous values.  Per-pixel regression has been
used for tasks such as predicting depth and surface normals~\cite{eigen2015predicting, kong2018recurrent}.  Our approach to mapping worm body coordinates was inspired by
the work of G{\"u}ler et. al. on ``DensePose'' which estimated dense correspondences
between image pixels and surface patches of a canonical 3D human body
model~\cite{alp2017densereg, alp2018densepose}.

\begin{figure}[t]
\centering
\includegraphics[width=\linewidth]{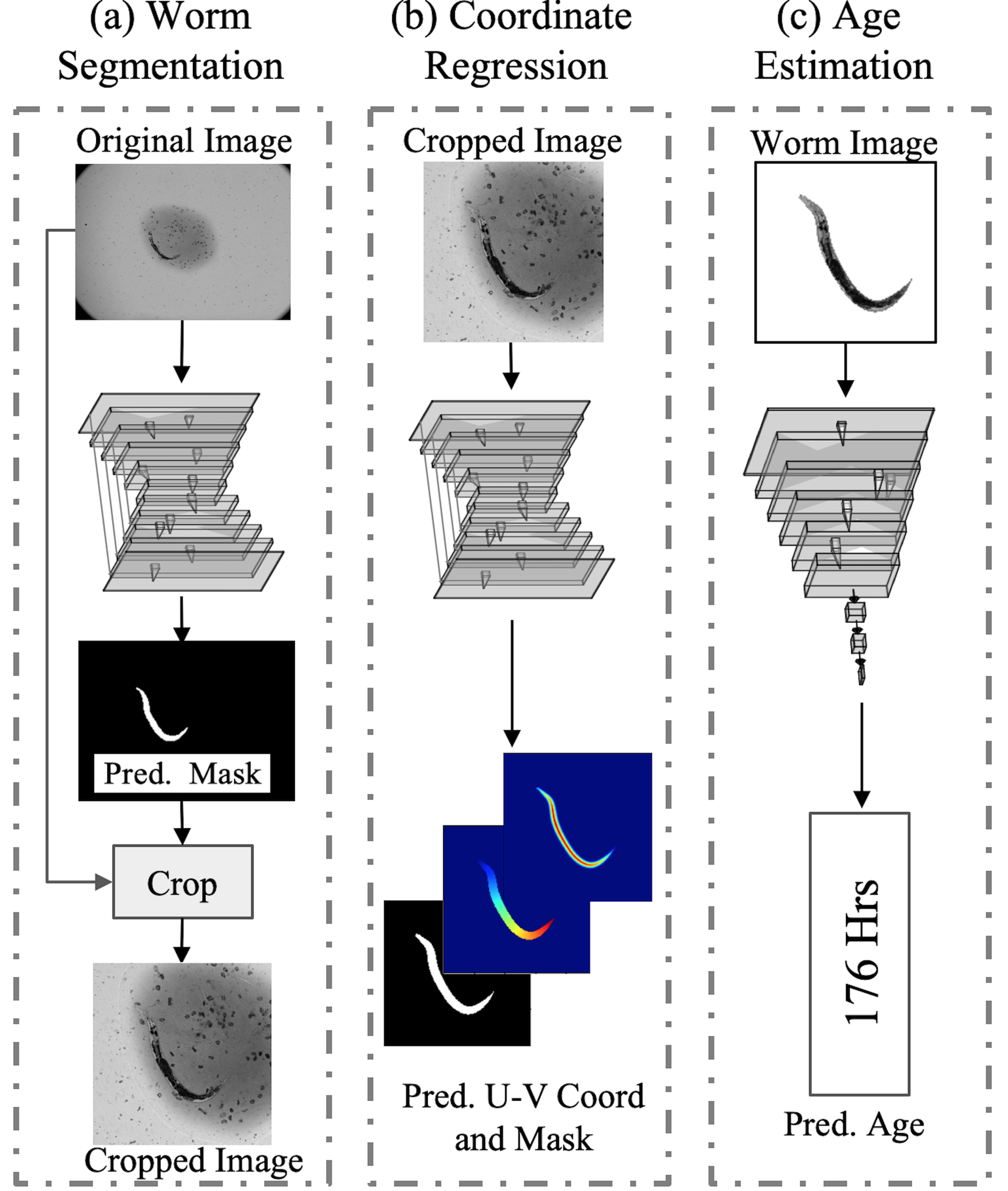}
%\vspace{-3mm}
   \caption{
   ``\emph{Celeganser}'' includes three models:
   (a) worm segmentation at coarse scale, (b) worm body coordinate regression and (c) age estimation.
   Using a downsized input image, model (a) predicts a binary
   segmentation of the worm region. This is used to localize and crop the worm region
   from the original full-resolution data.
   The cropped image is then fed to the model (b) for boy coordinate regression and fine segmentation. The segmented worm is the input to the third model (c) for age estimation.
   }
%\vspace{-3mm}
\label{fig:flowcharts}
\end{figure}

\section{\emph{Celeganser}: C. elegans Analyzer}
\label{sec:celeganser}

Our system consists of
three independently trained models for
worm segmentation,
worm body coordinate regression
and worm age estimation, respectively.
Figure~\ref{fig:flowcharts} shows the flowcharts of
the three modules, respectively.
Note that we train them sequentially,
and they also work in a sequential order in the system,
as the output of the previous module is the input to the current one.
In this section,
we elaborate the three models in the same order.

%We also discuss in-depth how to represent worm body coordinate (UV-space)
%for better learning.

\subsection{Worm Coarse Segmentation and Localization}
In our work, the microscope produces high-resolution images ($2160\times 2560$) of
individual worms captured once every $\sim$3.5 hours over the two-week lifespan of the worm.
Feeding such a high-resolution image into the CNN module directly for any final predictions
is unnecessarily costly in memory and computation.
Therefore,
we adopt a coarse-to-fine strategy.
To localize the worm we take as input a downsized image
(bilinear interpolated into $512\times 512$)
and perform coarse segmentation.
The output is a binary mask (after thresholding) of the same size $512\times 512$.
Based on the binary output mask,
we crop the raw image for the worm at original scale,
and obtain a $960\times 960$ sub-image capturing the worm.
For our data imaged at 5-10x magnification, we found that
a $960\times 960$ sub-image is sufficient to encapsulate any single worm.
The cropped sub-image serves as input to the second module for fine-grained prediction.

%Coarse segmentation model is a U-net based model which is also used to pre-process the image data. The C. elegans image sizes and resolution are too high and only center part of the huge image is the area where the worm is active. So, to reduce the amount of calculations and avoid unnecessary calculations, we need to extract the C. elegans in the images.

Predicting a segmentation for the worm can be treated as a two-class semantic segmentation
problem (worm foreground vs. background). Using CNNs for semantic segmentation is an active
area of research, and many excellent approaches are available in
literature~\cite{long2015fully, chen2017deeplab, ronneberger2015u}.
We adopt the U-Net architecture~\cite{ronneberger2015u},
consisting an encoder based on ResNet34~\cite{he2016deep} as the backbone,
and a decoder which has skip connections to the encoder and upsampling layers for the single mask
output at the input resolution.
Figure~\ref{fig:flowcharts} (a) depicts the flowchart of this model.

For training this model,
we use a per-pixel binary cross entropy loss.
We also insert the loss at multiple scales ($S=5$ scales in total)
at each skip connection layer,
weighted by the resolution of feature activations.
The loss for a single image is the sum of binary
cross entropy over all scales:
\begin{equation*}
    L_{seg} =-\sum_{s=1}^S \frac{1}{\vert {\cal I}_s \vert} \sum_{i \in {\cal I}_{s}}
    \big( y_i\log x_i +(1-y_i)\log(1-x_i)\big)
\label{eq:segm}
\end{equation*}
where $x_i$ is the predicted probability for pixel $i$ being worm and
${\cal I}_s$ is the set of pixel predictions at scale $s$.
$y_i$ is the ground-truth label, indicating whether the
pixel $i$ belongs to worm foreground ($y_i=1$)
or background ($y_i=0$).

% Once training the model,
% we use it to output a raw segmentation mask at a downsized size.
% Based on the segmented worm individual,
% we find the corresponding region on the raw image ($2160\times 2560$),
% and crop a $960\times 960$ sub-image for later use.

%
%For Coarse Segmentation model, we used U-net based model with skip connection to compute multi-scale loss and in each scale, we calculated the binary cross entropy loss by pixel-wise sigmoid over the output map. We calculated multi-scale loss($S$ scale in total) and in each scale, we applied the binary cross entropy loss at each pixel position $x \in \mathbb{Z}_i^2$. $p_m(x)$ is the class at $x$ position in ground truth mask and $p_o(x)$ is the output at $x$ position.
%\begin{equation}
%    L_{seg} =-\sum_{i = 1}^{S} \sum_{x\in \mathbb{Z}_i^2} p_m(x)log(p_o(x)) + (1-p_m(x))log(1- p_o(x))
%\end{equation}

\subsection{Body Coordinate Regression}
The second model, as shown by Figure~\ref{fig:flowcharts} (b),
takes as input the previous cropped sub-image
and outputs a fine segmentation mask and worm body coordinate prediction.
As we mentioned earlier,
our hypothesis with this model is that,
once it can regress the worm body well,
it should also capture meaningful information from internal structure of
the worm individual.
In this subsection,
we present how we represent the worm body coordinate for learning to regress.

\subsubsection{Worm Body Coordinate Representation}

\begin{figure}[t]
\begin{center}
\includegraphics[width=1\linewidth]{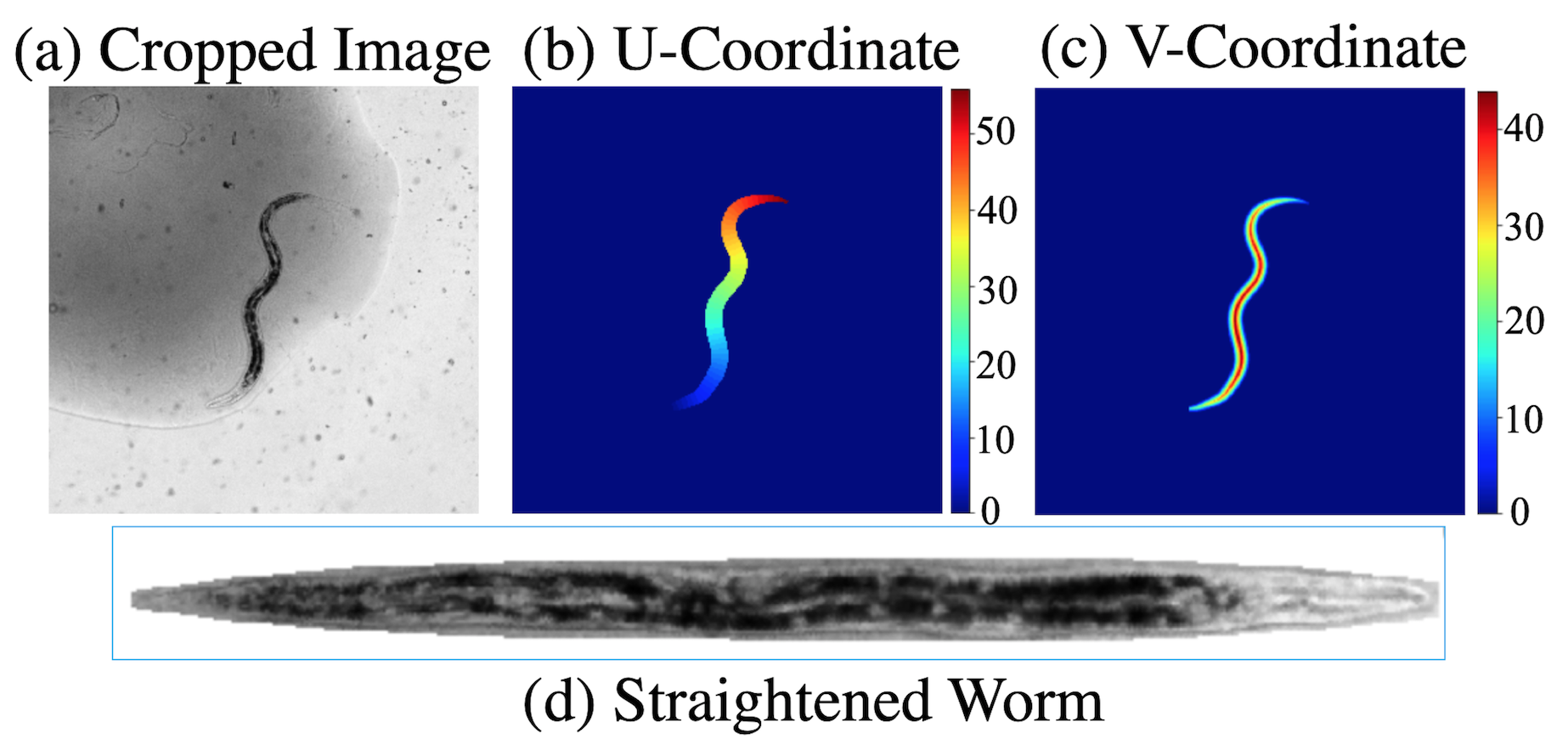}
\end{center}
   \caption{
   For a given cropped image (a),
   we train for regressing towards its worm UV-coordinate on the body pixels, as shown by (b) and (c) respectively.
   Based on the predicted coordinates,
   we can straighten the worm according to a defined ``canonical'' shape,
   as shown in (d).
   This helps us analyze worm age and thus the life/health-span in later work.
   More straightened worms can be found in Figure~\ref{fig:whole_result} with both ground-truth and predicted UV-coordinate.
   }
\label{fig:coordinate_straighten}
\end{figure}

\begin{figure}[t]
\begin{center}
\includegraphics[width=1\linewidth]{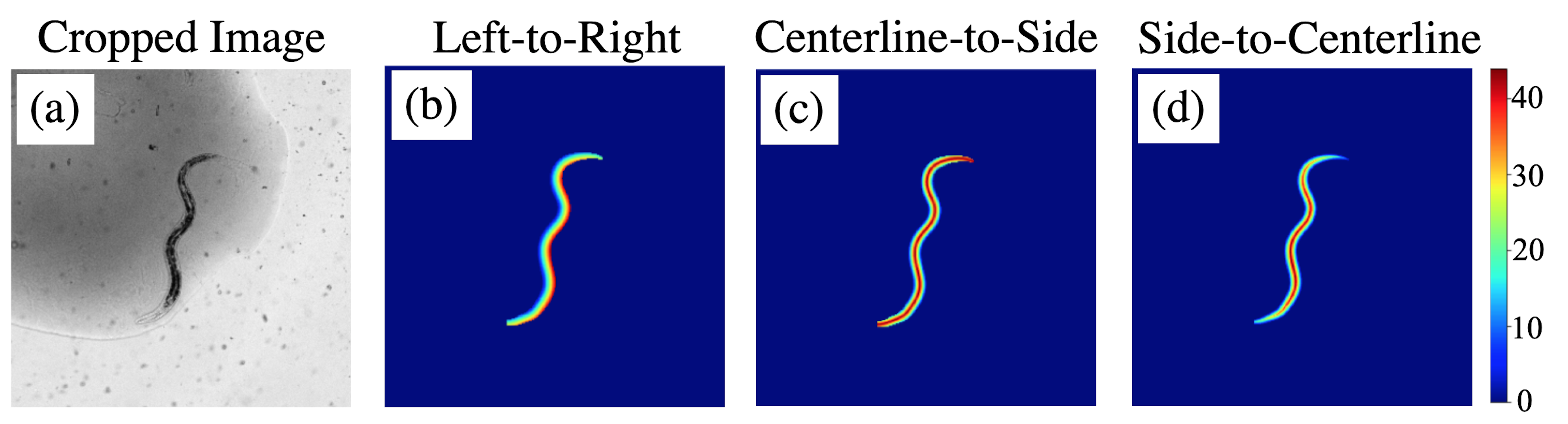}
\end{center}
   \caption{
   We consider three different representation of the V-coordinate for a given image (a):
   (b) left-to-right, (c) centerline-to-side and (d) side-to-centerline.
   We analyze them in depth in the main text,
   leading to our final decision to adopt (d) for more sensible representation.
   }
\label{fig:coordinate_representation}
\end{figure}

We would like to define a canonical coordinate system for the worm body which
we generically refer to as the UV-coordinate system (in contrast to the XY-coordinate
system describing pixel coordinates in the image).
A natural choice is to consider one coordinate that specifies the location along
the anterior-posterior axis (from head to tail) and a second to be the
dorsal-ventral axis (from back to belly) as worms typically ``swim'' on their
sides as viewed through the microscope. Given these axes, there remains
flexibility in exactly what coordinates to use and how they are scaled.

In this paper, we take U-coordinate as the the distance along the worm centerline
(in pixels or percent-body-length), from tail to head. For the V-coordinate we considered
three different representations which are visualized in Figure~\ref{fig:coordinate_representation}.
While it is generally straightforward to translate between these representations, in practice we
found that the choice of representation effects learning and prediction accuracy.

\noindent{\bf Left-to-Right Representation} as shown in Figure~\ref{fig:coordinate_representation} (a) is the most straightforward approach. Similar to the U-coordinate, we could take the V-coordinate to run left-to-right orthogonal to the centerline and range from zero up to
the width. However, we find that such a left-to-right V-coordinate is a difficult regression
target since the value is large on the right edge and then changes to 0 on the background.
Such a sharp change can make learning struggle on the right, causing problematic artifacts during inference. Moreover, we note that in the lab environment, worms can roll over which makes identifying the true left (ventral) and right (dorsal) difficult, even for human annotators.

\noindent{\bf Centerline-to-side Representation} is another we consider
where the centerline has a fixed maximum coordinate value (e.g., the maximum observed
width), and the V-coordinate decreases in proportion to the distance from the centerline.
Figure~\ref{fig:coordinate_representation} (b) shows a visualization this representation.
It remedies the sharp change on the worm body edge and avoids dorsal/ventral ambiguity.
However, such a representation tends to result in artifacts near the head and tail where
the width goes to zero.

% in straightening the worm to a canonical shape.\footnote{Straightening
% helps investigate correlation between internal structure of the worm and its health status.
% However, it is beyond our scope to study straightened worms for correlation between worm internal structure and its age (life/health-span).}
% Because this makes the straightened worm too much rectangular,
% distorting a lot on the head and tail parts.

\noindent{\bf Side-to-Centerline Representation}.
We found the best V-coordinate representation is to utilize the distance from the
side as depicted in Figure~\ref{fig:coordinate_representation} (c).
``Sides-to-Centerline'' means that, instead of starting from centerline,
the V-coordinate indicates the distance to the (nearest) boundary orthogonal
to the centerline. This is similar to a distance transform and has the appealing
properties that the boundary has a U-value of 0 and the maximum value, located
at the centerline indicates the width of the body at that V-coordinate.

\noindent{\bf Straightening}. As an alternative to visualizing the predicted
UV-coordinates in the original image plane, we can ``straighten'' the worm by
using the predicted coordinates to warp the worm image into a standardized
coordinate system. Figure~\ref{fig:coordinate_straighten} shows an example of
such a visualization where the bottom panel shows brightness values from the
original image displayed in a canonical pose. When performing this mapping for
the centerline-to-side representation, we disambiguate the V-coordinates for
the left and right sides based on the predicted head-tail orientation.

% Once again, studying in-depth the internal structure of individual worm,
% or its life/health-span, with the straightened worm is out of the scope of this paper.
% We expect to reveal more as future work.

\subsubsection{Worm Body Coordinate Regression}
Similar to worm segmentation module,
we also turn to a U-Net architecture for dense regression at pixel level.
The main task is to regress every pixel into UV coordinates.
However,
we note that the worm body is what we really care about, but not the background.
Therefore,
we also train this module to output a fine segmentation mask indicating where
the UV-coordinates are valid.

For learning the segmentation mask output,
we use the same binary cross entropy loss as Eq.~\ref{eq:segm} to train for segmentation,
as denoted by $L_{seg}^{fine}$ (fine segmentation as opposed to coarse segmentation
in the first module).
For UV-coordinate regression,
we simple adopt the L1 loss as below (we omit notation {\tt sum-over-images} for brevity):
\begin{equation}
    L_U =\sum_{s=1, i\in {\cal I}_s}^{S}  m^s_i \cdot | u^s_i - {\bar u}^s_i |,
\end{equation}
\begin{equation}
    L_V =\sum_{s=1, i\in {\cal I}_s}^{S}  m^s_i \cdot | v^s_i - {\bar v}^s_i |,
\end{equation}
where $m^s_i$ is output (after sigmoid transformation of the logits)
from the current segmentation output $\M^s$ at scale $s$.

We note that an alternative would be to mask the loss using the ground-truth
segmentation. Interestingly, we found that masking with the predicted mask yielded
slightly better models. We conjecture the reason is that the predicted segmentation
mask aligns better with the worm shape and allows the model to ignore some ``hard''
pixels that are included in the ground-truth segmentation mask.

In summary,
the total loss to train the body coordinate regression module is the following:
\begin{equation}
    L_{reg} = L_U + L_V + L_{seg}^{fine}
\end{equation}

\subsection{Worm Age Estimation}

With similar motivation as in worm body coordinate regression,
we hypothesize that if a model is able to predict worm age,
it must learn to leverage internal structure of the worm individual
to some extent. Such learned knowledge will in turn help us investigate
worm health status.

Worm age estimation is essentially a regression problem,
regressing the input image into a continuous value indicating the predicted age (in hours).
Therefore, we build upon a simple feed-forward ResNet34 network~\cite{he2016deep},
modifying its last layer to output a single continuous value.
The same backbone enables us to study how pre-training
helps improve age estimation,
either fine-tuning from ImageNet-pretrained model,
or the one trained for body coordinate regression.
Similar to body coordinate regression,
we simply adopt the L1 loss summing over all the $N$ images,
the ground-truth age in minutes $a_j$ and the predicted age
$a'_j$ for a specific image $j$:
\begin{equation}
    L_{age} = \sum_{j=1}^N  |a_j - a'_j|.
\end{equation}
We do not re-weight pixel-level loss w.r.t ground-truth ages.
We find that the simple L1 loss suffices,
introducing little biased prediction as demonstrated in experiments
on the ground-truth and prediction distribution over ages.

% \begin{figure*}[t]
% \begin{center}
% \includegraphics[width=\linewidth]{latex/Result/worm display.png}
% \end{center}
%   \caption{Example of cropped image of C. elegans after Coarse segmentation model and cropping operation under 5X magnification microscope.}
% \label{fig:worm_display}
% \end{figure*}

\section{Experiment}
\label{sec:exp}

% We carry out extensive experiments to validate our \emph{Celeganser} system,
% through studying all the three models independently.
We start by describing our dataset and our implementations,
before diving into each model and validating performance of each.

\begin{figure}[t]
\begin{center}
\includegraphics[width=0.995\linewidth]{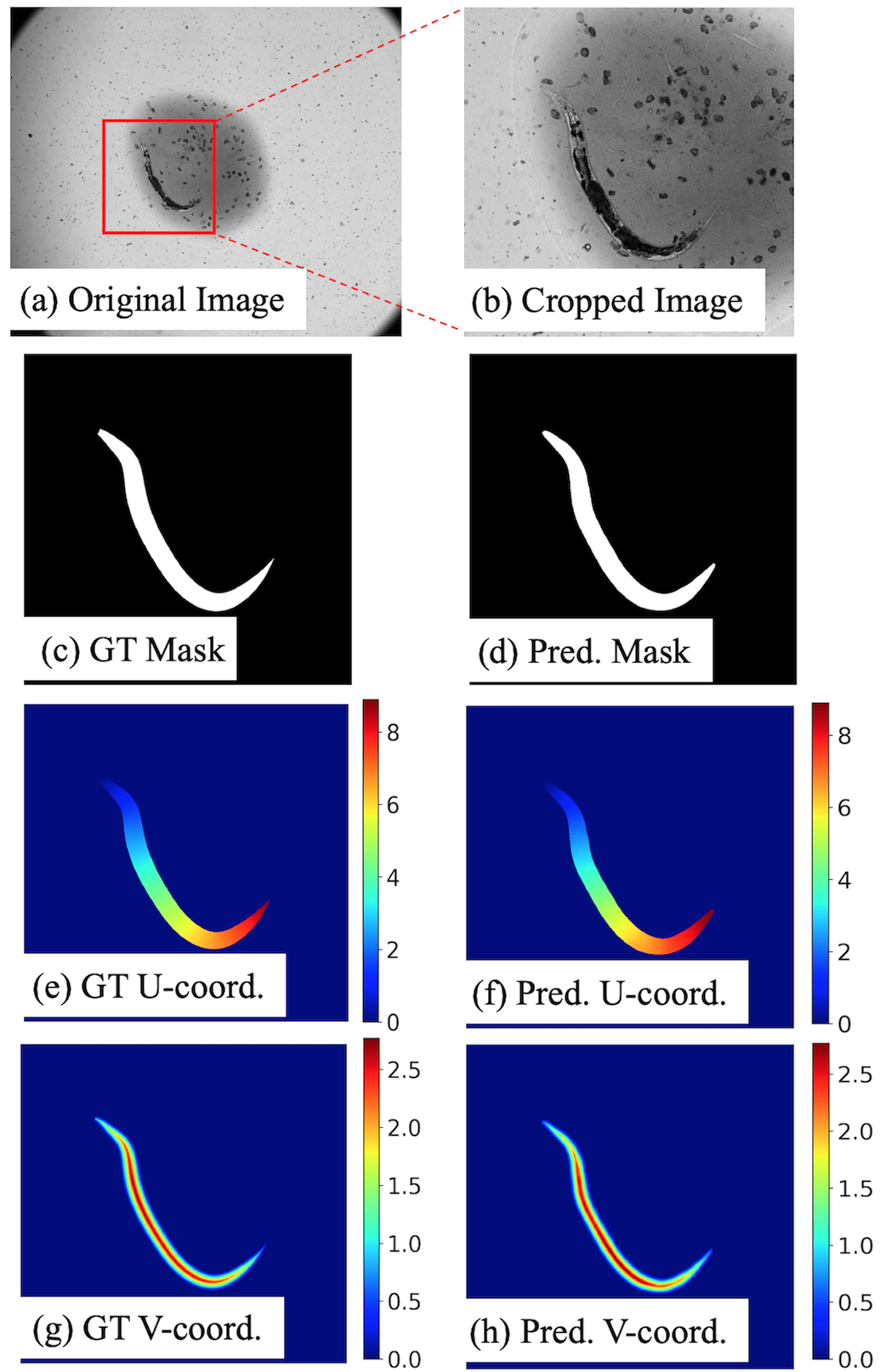}
\end{center}
   \caption{Worm body coordinate regression model takes as input the sub-image (b) which is cropped from the original image (a),
   and outputs worm segmentation mask (d) and UV coordinate predictions (f) and (h), respectively.
 }
\label{fig:reg_result}
\end{figure}

\subsection{Implementation Details}

\noindent\textbf{Dataset}.
%\shu{Zach may write how to collect dataset from here?}
Individual worms were imaged at 5x or 7x resolution at regular intervals ($\sim$3.5hours) from larval stage to death, yielding a final dataset of 10,075 16-bit high resolution images, each of which contains a single worm.  To annotate UV coordinates we leveraged
the zlab toolbox,\footnote{https://github.com/zplab/elegant} which allows the user to draw the
centerline as a spline curve and indicate other keypoints corresponding to organs and ground-truth
segmentation mask. Annotation and curation was performed by a group experienced biologists

% We trained and evaluated the models on the C. elegans dataset under 5X magnification microscope including 10273 16-bit high resolution images of xxxx C. elegans and their human-annotated poses information. Each C. elegans was annotated its own center line, width at each center line sampling points and its anatomical keypoints, which can be used to U-V coordinates in any representation format. Also, there are additional information for each C. elegans, including length, max-width, age(post-hatch in hours), stage(larval or adult) and self-overlap-fraction. C. elegans was captured by under microscope from early larval stage to death.
Over our annotated worm dataset, we split it into training (9,056 images) and
validation (1,019 images). Our split was performed w.r.t worm identities so that
a given individual worm never appears in both training and test.

\noindent\textbf{Network Design}.
We use ResNet34~\cite{he2016deep} as the encoder/backbone for all our three models.
We observe that, given the amount of training data available, deeper models do not
necessarily improve performance further.
For per-pixel prediction by worm segmentation and body coordinate regression,
we build a decoder with skip connection to the encoder with same structure,
consisting upsampling layers, convolution layers,
ReLU layers~\cite{nair2010rectified} and Batch Normalization layers~\cite{ioffe2015batch}.

\noindent\textbf{Training}.
We train each model individually for 100 epochs using the Adam optimizer~\cite{kingma2014adam},
with initial learning rate 0.0005 and coefficients 0.9 and 0.999
for computing running averages of gradient and its square.
We decrease learning rate by half every 20 epochs.
%with initial learning 5e-4 and 0.5 decay rate for each 20 epochs.
For age estimation model,
we fine-tune a pre-trained checkpoint (e.g., the
body-coordinate regression model)
with new decoder initialized randomly.
%unchanged and only fine-tuned the pooling and fully connected layers.
We train our model using PyTorch~\cite{paszke2017automatic}
on a single NVIDIA TITAN X GPU.

\subsection{Worm Segmentation}

{
\setlength{\tabcolsep}{1.2em} % for the horizontal padding
\begin{table}[t]
\centering
\caption{Comparison of worm segmentation by the
coarse segmentation module and the body coordinate regression module.
We report Intersection-over-Union (IoU) and pixel classification accuracy (Acc.). Larger numbers mean better performance $\uparrow$.}
\begin{tabular}{l|c|c}
\hline
    & {\tt Coarse Segm.}  & {\tt Fine Segm.} \\
\hline
IoU  &  0.8482  & {\bf 0.9033} \\
\hline
Acc. &  0.9945  & {\bf 0.9964} \\
\hline
\end{tabular}
\label{tab:seg_evaluation}
\end{table}
}

\begin{figure*}[t]
\centering
\includegraphics[width=0.9995\linewidth]{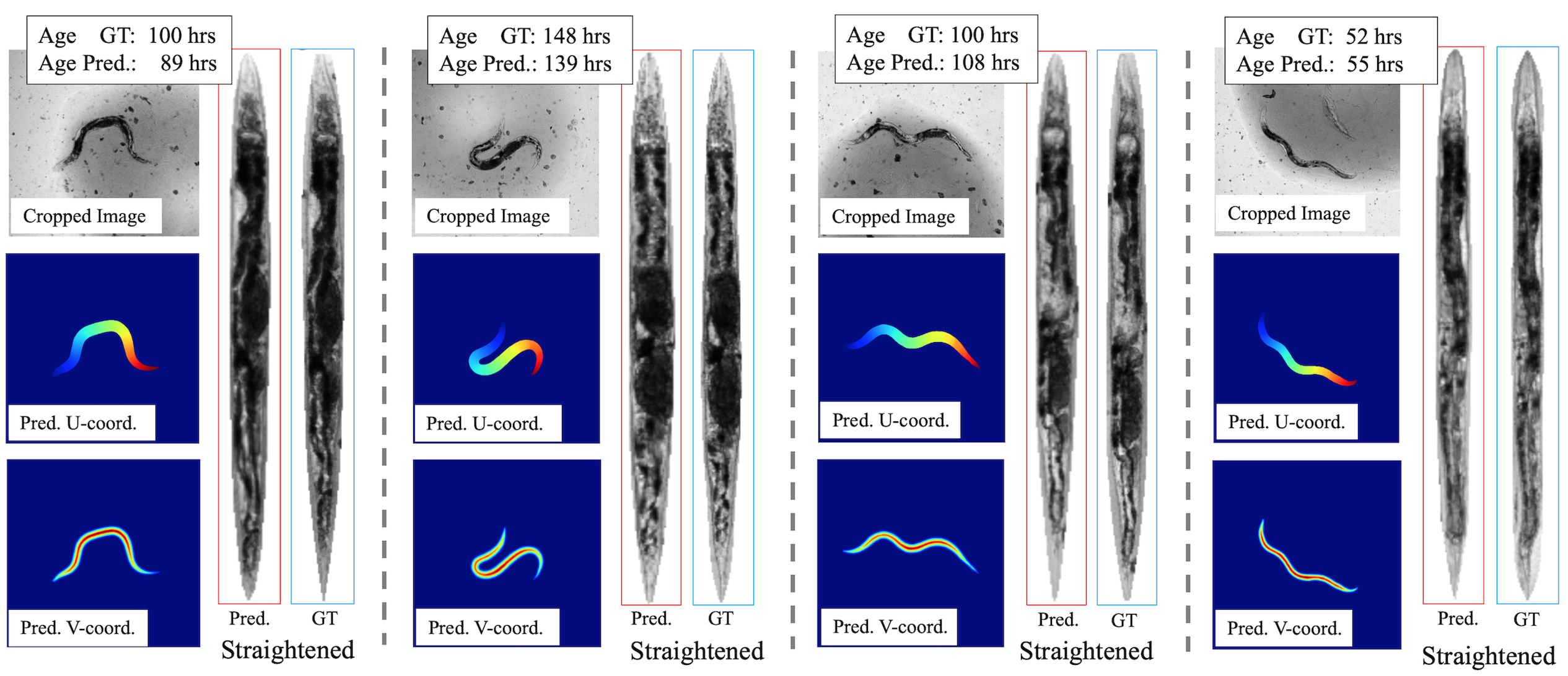}
   \caption{
   We visualize the body UV-coordinate regression output.
   With the ground-truth and predicted UV-coordinates,
   we straighten the worm into ``canonical shape'', respectively.
   Visually we can see the straightened worms match quite well
   between using predicted and ground-truth UV's.
   }
\label{fig:whole_result}
\end{figure*}

We evaluate the accuracy of the segmentation masks predicted by both the
coarse (low-res) segmentation model and the fine (high-res) mask.
% Training to segment worm in our data is simple,
% and measuring the segmentation performance is also straightforward.
% Therefore,
% we compare between the first two models in our Celeganser system,
% as both the first two models output worm segmentation masks.
Note that the segmentation masks from the two models have different resolution,
due to both resizing and cropping.
For fair comparison,
we resize the coarse segmentation mask (sigmoid transformed map)
back to the original image size (using bilinear interpolation),
binarize the rescaled map, and crop it to match the high-res sub-image.
Then we have two segmentation masks that have the same size and same region.
We report the performance measure w.r.t two metrics,
mean Intersection-over-Union (IoU) over all the testing images,
and mean accuracy over all pixels of all testing images.

Table~\ref{tab:seg_evaluation} lists the comparison,
showing that the second model performs much better than the first one in terms
of learning to segment worms.
This is because of two reasons.
First, high-resolution input image as the input to the second model
provides with finer-grained pixel information,
which is helpful for predicting masks; whereas upsampling the coarse segmentation output
may introduce artifacts that harm segmentation performance.
Second,
the first model works on smaller worms compared to the input image (biased towards the background),
while the second model receives cropped image as input and thus works on a much balanced binary
classification/segmentation problem.
While the first model merely outputs a coarse segmentation mask,
we note it is sufficient for us to crop the worm region based on the imperfect prediction,
and move forward with fine segmentation and worm body regression.
Figure~\ref{fig:reg_result} (d) shows the segmentation mask by the second model.

\subsection{Worm Body Coordinate Regression}

{
\setlength{\tabcolsep}{1.0em} % for the horizontal padding
\begin{table}[t]
\centering
\caption{Average absolute error over for body coordinate regression, with or without masking in the loss function. We use the predicted segmentation mask to mask off background pixels.
To compute Avg. Abs. Err., we average over both U and V predictions within
the worm region specified by the ground-truth worm mask.
Smaller error is better $\downarrow$.}
\begin{tabular}{l|c|c}
\hline
                & {\tt w/o mask} &  {\tt w/ mask} \\
\hline
Avg. Abs. Err.  & 0.3361    & {\bf 0.2821} \\
\hline
\end{tabular}
\label{tab:reg_evaluation}
\end{table}
}

For worm body coordinate regression,
we measure the averaged L1 difference over all pixels between ground-truth UV coordinates
and the predictions.
However, as discussed, we have an option in training of whether to use the (predicted)
segmentation mask to mask off background when computing the regression loss.
Therefore,
we compare two models that are trained without and with such a masking mechanism.
Table~\ref{tab:reg_evaluation} lists the comparison.

We can clearly see from Table~\ref{tab:reg_evaluation}
that with the masking mechanism during training,
the model outperforms the counterpart (without masking) significantly.
This demonstrates the benefit of enforcing the model to focus on the worm
region, instead of all pixels of the input image.
Figure~\ref{fig:reg_result} shows
one example of predicted UV-coordinate and the fine segmentation mask,
compared with the ground-truth;
while Figure~\ref{fig:whole_result} shows more UV coordinate regression results.

% We aim to study individual worms,
% and each image contains only one worm.
% This makes our data simpler than complicated computer vision benchmarks that
% focus on natural images.
% Therefore,
% it is not surprising that we can train relatively easily for worm
% segmentation and body coordinate regression.
It is worth noting that,
with the predicted UV-coordinate,
we can straighten the worms into a ``canonical shape'',
allowing for better analysis w.r.t internal structures, e.g.,
organs' location and size.
Figure~\ref{fig:whole_result}  qualitatively compares
the straightened worms by ground-truth UV and the predicted UV coordinates,
respectively.
We can see straightened worm are fairly comparable,
enabling us to find specific organs to compare the size and location relative
to the worm body.

\subsection{Worm Age Estimation}

{
\setlength{\tabcolsep}{0.55em} % for the horizontal padding
\begin{table*}[t]
\centering
\caption{Age estimation performance measured by averaged absolute difference (over all
 validation images)
 with the ground-truth age (in hours).
 We compare three models (rows), training from scratch, fine-tuning from ImageNet-pretrained checkpoint and fine-tuning from our coordinate regression model.
 We also compare five different input format as listed in the five columns.
 Refer to the main text for detailed analysis.
}
\begin{tabular}{l|c|c|c|c|c}
\hline
 & {\tt Raw~Image} & {\tt Worm~Only} & {\tt Background~(BG)} & {\tt Silhouette} & {\tt Silhouette+BG}  \\
\hline
training-from-scratch & 18.50  & 31.08  & 36.03 & 69.23 & 18.51  \\
\hline
ImageNet-pretrain & 17.63 & 17.94  & 33.35 & 65.89 & 18.06   \\
\hline
coord.-reg.-pretrain  & 14.50 & 20.90   & 26.43 & 65.08 & 17.00 \\
\hline
\end{tabular}
\label{tab:ablatoin_age_est}
\end{table*}
}

\begin{figure}[t]
\centering
\includegraphics[width=1.0\linewidth]{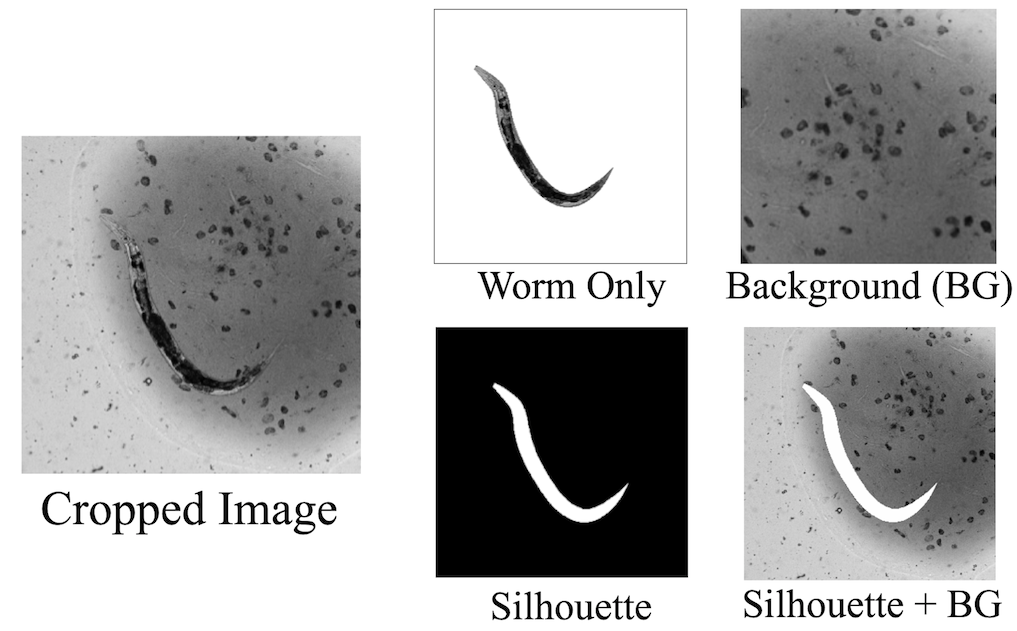}
   \caption{
   Feature masking study in which we train models on different input modalities for worm age
   estimation. This helps disentangle which features are informative for age-estimation: the background environment, the worm size and shape (silhouette), or the internal structure of the worm individual.  For example, even without looking at the worm it is possible to tell something
   about the age since the waste visible in the background increases over time.
   }
\label{fig:worm_decompose}
\end{figure}

We hypothesize that a model which is good at body-coordinate regression will likely
extract features of internal texture that are also useful for predicting age. We
evaluate this using a pre-training transfer experiment.  First we randomly initialize
the weights to train from scratch. This provides a ``lower bound'' on the performance.
Second, we follow a popular practice that we fine-tune from ImageNet pretrained checkpoint~\cite{deng2009imagenet} for age estimation.
Pre-training provides a better staring point than randomly initialized weights
for targeted tasks. Third,
we fine-tune our worm body coordinate regression model,
which has already seen worm data and we expect such pre-training to perform better
for age estimation.

\noindent{\bf Feature Masking}
A key challenge in drawing conclusions from powerful supervised machine learning models
is that they easily latch on to image features which may be independent of the biology
of interest but happen to be correlated. For predicting age, this could include factors
such as the presence of tracks and waste in the growth media, condensation, drift in
focus or illumination over time, etc.  To control for these factors, we also
evaluated how age-estimation accuracy varies when we mask aspects of the input image.

In addition to reporting the worm estimation performance with the input as the cropped {\tt Raw Image}, we consider the other four input formats, as shown in Figure~\ref{fig:worm_decompose}.
{\tt Worm Only} means that we use the worm segmentation output to mask off
all the background pixels, then we input this single worm to train for age estimation.
This helps us understand how good age estimation we could achieve merely depending on features of the worm.
{\tt Background} (BG) is a random region cropped from the raw image but excluding any pixels belonging to the worm.
We feed the background region to train for age estimation, with the real ground-truth age of
the worm living in that environment.
Note that background accumulates worm's waste, eggs and and tracks during time.
Therefore the visual properties of the background are actually a good indicator of
how long the worm has lived in the environment without seeing the animal itself.
We consider an even more challenging setup,
using {\tt silhouette} and {\tt silouette+GB} to train for age estimation.
The former helps show how much the worm shape and size correlate with its (estimated) age,
while the latter will demonstrate how we can achieve with the combination of
background information and shape.

\begin{figure}[t]
\centering
\includegraphics[width=0.9997\linewidth]{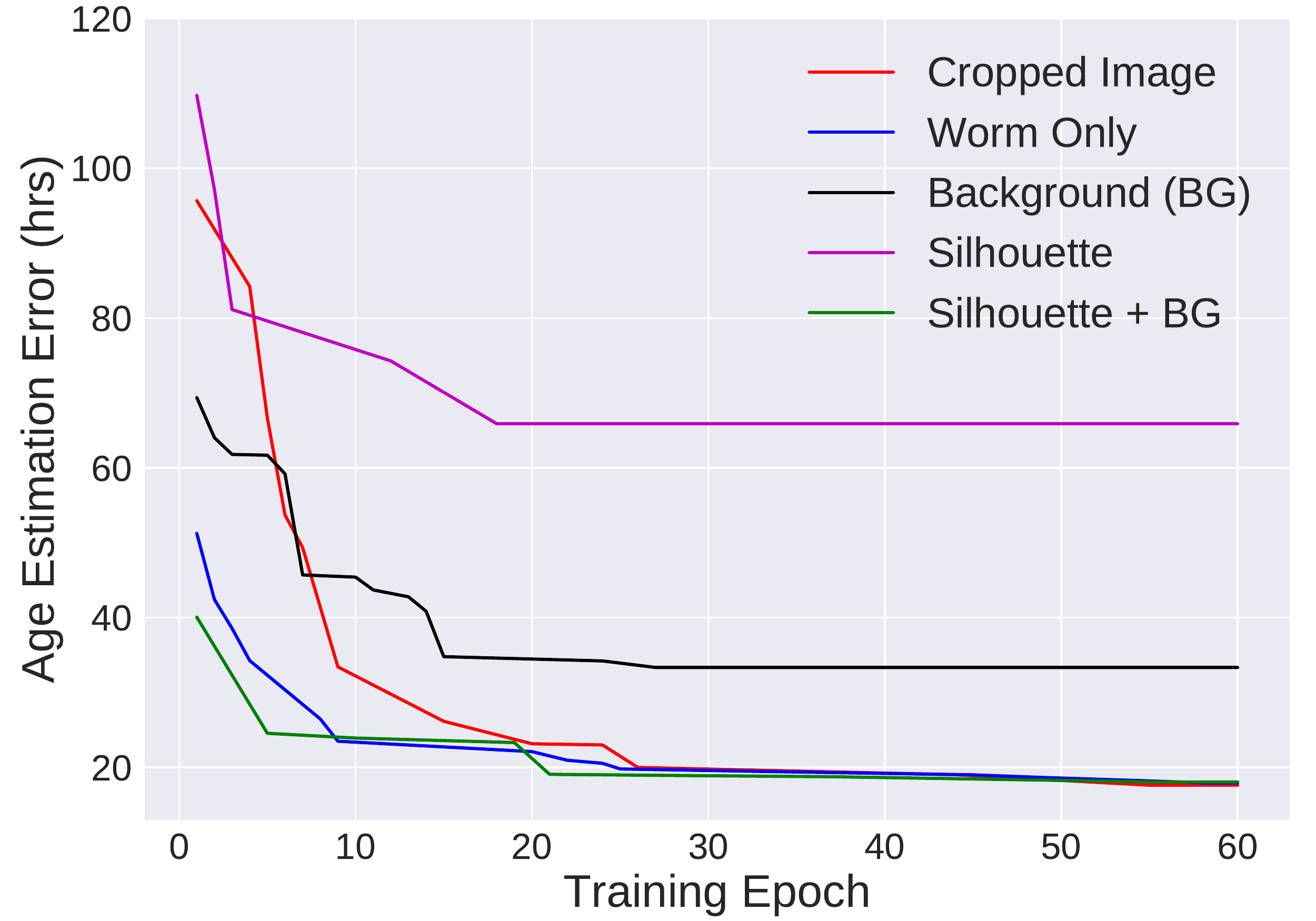}
   \caption{We plot the age estimation error in the validation set in the first 60 epochs. The models chosen here are  the ones fine-tuned from an ImageNet-pretrained checkpoint, with different input formats.
   We can see all models converge at the similar speed,
   and slowly keep decreasing over time.
   Moreover, none of the models overfit (or get worse) even being trained long enough.
   }
\label{fig:age_loss}
\end{figure}

Figure~\ref{fig:age_loss} plots age estimation  error on the validation set with the models trained by fine-tuning the ImageNet pretrained checkpoint.
We can see the models, even with different input formats,
converge smoothly at the similar speed.
Moreover, they do not overfit training data (or become worse on validation data), probably owing to the use of Batch Normalization in the models.

Table~\ref{tab:ablatoin_age_est} lists detailed performance comparisons all the three models
and five different input modalities. We note several interesting observations:
\begin{itemize}
  \item Background indeed provides much information for age estimation, as we expected.
    The model trained on background only even outperforms the one trained on {\tt silhouette}.
    This indicates the worm shape alone provides limited information for age estimation.
  \item Silhouette+BG shows much better performance than {\tt Background} or {\tt Silhouette}.
    This demonstrates the joint force by background and worm shape provides more informative
    cue for age estimation. However, we note that the predictive value of the background is
    entirely an artifact of the experimental condition (starting with a clean environment)
    and thus likely has very little to do with the biology of aging itself\footnote{Of course
    the availability of food and other environmental features and stressors certainly do affect aging but in the context of the present experiment they are not controlled for independent
    of age and hence should be ignored.}
  \item Using model pre-training outperforms substantially the model trained from scratch.
    Fine-tuning our coordinate regression model is even better than using the ImageNet-pretrained model with raw images as input, supporting our hypothesis that pretraining for body-coordinate
    regression results in useful features that capture the shape and content of the worm.
  \item Masking out the internal content of the worm (Silhouette+BG) decreases performance
    relative to the Raw Image, but the difference is small except for the coordinate-regression
    pretrained model which shows a much bigger benefit of adding the worm content (17.0 to 14.5)
  \item While using raw image consistently gets the best predictive performance for each model, we  note that {\tt Worm Only} varies a lot for different models: ImageNet-pretrain works the best,
  outperforming coord.-re.-pretrain, whereas training-from-scratch does not show competitive results (which is understandable). We believe the reason is that, while coordinate regression model,
  as the pretrained checkpoint, has seen the worm images,
  it actually does not see {\tt Worm Only} images which have very different distribution. However,
  the ImageNet-pretrained checkpoint offers a more generic feature extractor which makes itself more amendable for the {\tt Worm Only} images through fine-tuning.
\end{itemize}

\begin{figure}[t]
\centering
\includegraphics[width=\linewidth]{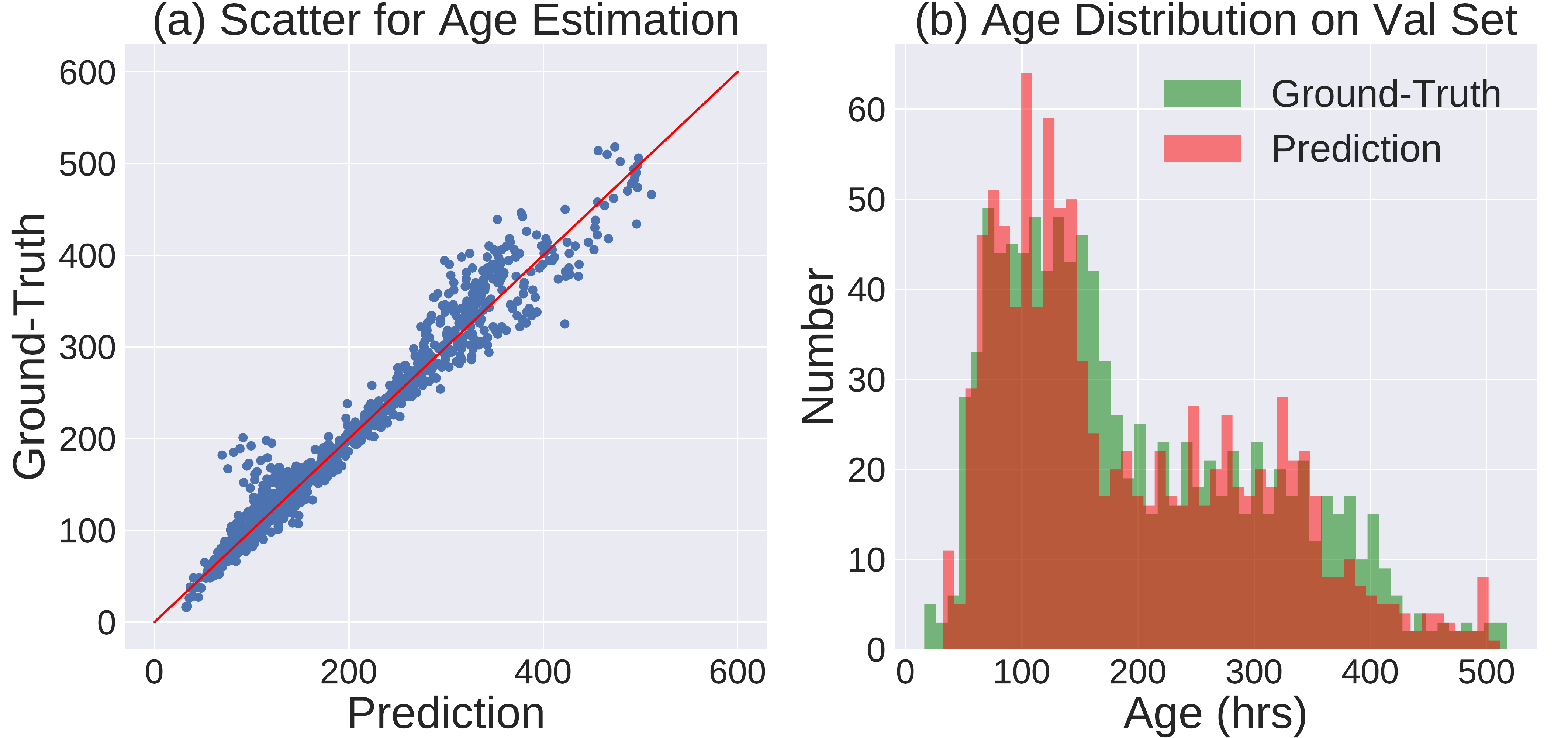}
%\vspace{-3mm}
   \caption{
   We demonstrate that the age prediction results align well with the ground-truth ages,
   based on two plots,
   (a) scatter plot of ground-truth vs. prediction of age estimation, and (b) histograms of ground-truth and prediction age in the validation set.
   }
\label{fig:age_distribution}
\end{figure}

\noindent{\bf Age Distribution matching}
We plot the distribution of ground-truth and age predictions in
Figure~\ref{fig:age_distribution},
(a) scatter plot to visualize the correlation between prediction and ground-truth,
and (b) overlaid histograms to understand how the predictions match the ground-truth
overall. We can see age prediction aligns quite well with the ground-truth. From the
scatter plot we can see the smallest errors are for juvenile worms ($<$100 hours)
and middle aged (180-300 hours) with some decrease in accuracy near the end of
life ($>$300 hours).

\section{Conclusion and Future Work}
\label{sec:conclusion}
We introduce ``\emph{Celeganser}'',
a prototyped automated system that assists analysis on single C. elegans worm
in terms of studying lifespan and healthspan.
Through extensive experiments,
we validate the modules in Celeganser, including semantic segmentation, body coordinate regression
and age estimation.
The system already achieves high-levels of accuracy and deploying this system will allow
for high-throughput analysis and provide interpretable and useful morphological features
for studying life/health-span. We hope our study also sheds light on how to represent
UV-coordinate and the importance of feature masking experiments in deriving biological
insights from machine learning models.

% With our hypothesis that the system, when able to regress body coordinate and estimate age well,
% should also capture correlation between worm internal organs and worm age (which is closely related to worm body length),
% and thus the life/health-span.
% Therefore,
% we expect to extend our prototyped system,
% and apply it to the research of life/health-span with C. elegans.

\begin{comment}
\begin{figure}[t]
\begin{center}
\includegraphics[width=1.0\linewidth]{Result/regression result -2.png}
\end{center}
   \caption{Result from fine-scale regression model}
\label{fig:reg_result-2}
\end{figure}

\begin{figure}[t]
\begin{center}
\includegraphics[width=\linewidth]{Result/regression result -3.png}
\end{center}
   \caption{Result from fine-scale regression model}
\label{fig:reg_result-3}
\end{figure}
\end{comment}

\section*{Acknowledgements}
The authors gratefully acknowledge Logan Tan, Nicolette Laird and Aditya Somisetty
of the Pincus Lab who created and curated the ground-truth image annotations.
This research was supported by NIH grant NIA R01AG057748, 
NSF grants IIS-1813785 and IIS-1618806, 
a research gift from Qualcomm, 
and a hardware donation from NVIDIA.
Shu Kong also acknowledges Kleist Endowed Fellowship for the generous support of inter-disciplinary research.
{\small
\bibliographystyle{ieee_fullname}
\bibliography{egbib}

\begin{thebibliography}{10}\itemsep=-1pt

\bibitem{alp2018densepose}
R{\i}za Alp~G{\"u}ler, Natalia Neverova, and Iasonas Kokkinos.
\newblock Densepose: Dense human pose estimation in the wild.
\newblock In {\em Proceedings of the IEEE Conference on Computer Vision and
  Pattern Recognition}, pages 7297--7306, 2018.

\bibitem{alp2017densereg}
Riza Alp~Guler, George Trigeorgis, Epameinondas Antonakos, Patrick Snape,
  Stefanos Zafeiriou, and Iasonas Kokkinos.
\newblock Densereg: Fully convolutional dense shape regression in-the-wild.
\newblock In {\em Proceedings of the IEEE Conference on Computer Vision and
  Pattern Recognition}, pages 6799--6808, 2017.

\bibitem{baeriswyl2010modulation}
Simon Baeriswyl, M{\'e}d{\'e}ric Diard, Thomas Mosser, Magali Leroy, Xavier
  Mani{\`e}re, Fran{\c{c}}ois Taddei, and Ivan Matic.
\newblock Modulation of aging profiles in isogenic populations of
  caenorhabditis elegans by bacteria causing different extrinsic mortality
  rates.
\newblock {\em Biogerontology}, 11(1):53, 2010.

\bibitem{brenner1974genetics}
Sydney Brenner.
\newblock The genetics of caenorhabditis elegans.
\newblock {\em Genetics}, 77(1):71--94, 1974.

\bibitem{chen2017deeplab}
Liang-Chieh Chen, George Papandreou, Iasonas Kokkinos, Kevin Murphy, and Alan~L
  Yuille.
\newblock Deeplab: Semantic image segmentation with deep convolutional nets,
  atrous convolution, and fully connected crfs.
\newblock {\em IEEE transactions on pattern analysis and machine intelligence},
  40(4):834--848, 2017.

\bibitem{cook2019whole}
Steven~J Cook, Travis~A Jarrell, Christopher~A Brittin, Yi Wang, Adam~E
  Bloniarz, Maksim~A Yakovlev, Ken~CQ Nguyen, Leo T-H Tang, Emily~A Bayer,
  Janet~S Duerr, et~al.
\newblock Whole-animal connectomes of both caenorhabditis elegans sexes.
\newblock {\em Nature}, 571(7763):63--71, 2019.

\bibitem{coudray2018classification}
Nicolas Coudray, Paolo~Santiago Ocampo, Theodore Sakellaropoulos, Navneet
  Narula, Matija Snuderl, David Feny{\"o}, Andre~L Moreira, Narges Razavian,
  and Aristotelis Tsirigos.
\newblock Classification and mutation prediction from non--small cell lung
  cancer histopathology images using deep learning.
\newblock {\em Nature medicine}, 24(10):1559--1567, 2018.

\bibitem{curtsinger1992demography}
James~W Curtsinger, Hidenori~H Fukui, David~R Townsend, and James~W Vaupel.
\newblock Demography of genotypes: failure of the limited life-span paradigm in
  drosophila melanogaster.
\newblock {\em Science}, 258(5081):461--463, 1992.

\bibitem{deng2009imagenet}
Jia Deng, Wei Dong, Richard Socher, Li-Jia Li, Kai Li, and Li Fei-Fei.
\newblock Imagenet: A large-scale hierarchical image database.
\newblock In {\em Proceedings of the IEEE Conference on Computer Vision and
  Pattern Recognition (CVPR)}, pages 248--255. Ieee, 2009.

\bibitem{eigen2015predicting}
David Eigen and Rob Fergus.
\newblock Predicting depth, surface normals and semantic labels with a common
  multi-scale convolutional architecture.
\newblock In {\em Proceedings of the IEEE international conference on computer
  vision}, pages 2650--2658, 2015.

\bibitem{esteva2017dermatologist}
Andre Esteva, Brett Kuprel, Roberto~A Novoa, Justin Ko, Susan~M Swetter,
  Helen~M Blau, and Sebastian Thrun.
\newblock Dermatologist-level classification of skin cancer with deep neural
  networks.
\newblock {\em Nature}, 542(7639):115--118, 2017.

\bibitem{gonczy2005asymmetric}
Pierre G{\"o}nczy and Lesilee~S Rose.
\newblock Asymmetric cell division and axis formation in the embryo.
\newblock {\em WormBook}, page~1, 2005.

\bibitem{he2016deep}
Kaiming He, Xiangyu Zhang, Shaoqing Ren, and Jian Sun.
\newblock Deep residual learning for image recognition.
\newblock In {\em Proceedings of the IEEE Conference on Computer Vision and
  Pattern Recognition (CVPR)}, pages 770--778, 2016.

\bibitem{ioffe2015batch}
Sergey Ioffe and Christian Szegedy.
\newblock Batch normalization: Accelerating deep network training by reducing
  internal covariate shift.
\newblock {\em arXiv preprint arXiv:1502.03167}, 2015.

\bibitem{johnson2001age}
Thomas~E Johnson, Deqing Wu, Patricia Tedesco, Shale Dames, and James~W Vaupel.
\newblock Age-specific demographic profiles of longevity mutants in
  caenorhabditis elegans show segmental effects.
\newblock {\em The Journals of Gerontology Series A: Biological Sciences and
  Medical Sciences}, 56(8):B331--B339, 2001.

\bibitem{kenyon1993c}
Cynthia Kenyon, Jean Chang, Erin Gensch, Adam Rudner, and Ramon Tabtiang.
\newblock A c. elegans mutant that lives twice as long as wild type.
\newblock {\em Nature}, 366(6454):461--464, 1993.

\bibitem{kenyon2010genetics}
Cynthia~J Kenyon.
\newblock The genetics of ageing.
\newblock {\em Nature}, 464(7288):504--512, 2010.

\bibitem{kimura1997daf}
Koutarou~D Kimura, Heidi~A Tissenbaum, Yanxia Liu, and Gary Ruvkun.
\newblock daf-2, an insulin receptor-like gene that regulates longevity and
  diapause in caenorhabditis elegans.
\newblock {\em Science}, 277(5328):942--946, 1997.

\bibitem{kingma2014adam}
Diederik~P Kingma and Jimmy Ba.
\newblock Adam: A method for stochastic optimization.
\newblock {\em arXiv preprint arXiv:1412.6980}, 2014.

\bibitem{kong2018recurrent}
Shu Kong and Charless~C Fowlkes.
\newblock Recurrent scene parsing with perspective understanding in the loop.
\newblock In {\em Proceedings of the IEEE Conference on Computer Vision and
  Pattern Recognition}, pages 956--965, 2018.

\bibitem{krizhevsky2012imagenet}
Alex Krizhevsky, Ilya Sutskever, and Geoffrey~E Hinton.
\newblock Imagenet classification with deep convolutional neural networks.
\newblock In {\em Advances in neural information processing systems}, pages
  1097--1105, 2012.

\bibitem{lecun1998gradient}
Yann LeCun, L{\'e}on Bottou, Yoshua Bengio, and Patrick Haffner.
\newblock Gradient-based learning applied to document recognition.
\newblock {\em Proceedings of the IEEE}, 86(11):2278--2324, 1998.

\bibitem{lin2020using}
Jiunn-Liang Lin, Wei-Liang Kuo, Yi-Hao Huang, Tai-Lang Jong, Ao-Lin Hsu, and
  Wen-Hsing Hsu.
\newblock Using convolutional neural networks to measure the physiological age
  of caenorhabditis elegans.
\newblock {\em IEEE/ACM Transactions on Computational Biology and
  Bioinformatics}, 2020.

\bibitem{long2015fully}
Jonathan Long, Evan Shelhamer, and Trevor Darrell.
\newblock Fully convolutional networks for semantic segmentation.
\newblock In {\em Proceedings of the IEEE conference on computer vision and
  pattern recognition}, pages 3431--3440, 2015.

\bibitem{mair2003demography}
William Mair, Patrick Goymer, Scott~D Pletcher, and Linda Partridge.
\newblock Demography of dietary restriction and death in drosophila.
\newblock {\em Science}, 301(5640):1731--1733, 2003.

\bibitem{nair2010rectified}
Vinod Nair and Geoffrey~E Hinton.
\newblock Rectified linear units improve restricted boltzmann machines.
\newblock In {\em Proceedings of the 27th international conference on machine
  learning (ICML-10)}, pages 807--814, 2010.

\bibitem{paszke2017automatic}
Adam Paszke, Sam Gross, Soumith Chintala, Gregory Chanan, Edward Yang, Zachary
  DeVito, Zeming Lin, Alban Desmaison, Luca Antiga, and Adam Lerer.
\newblock Automatic differentiation in pytorch.
\newblock 2017.

\bibitem{pincus2016ageing}
Zachary Pincus.
\newblock Ageing: A stretch in time.
\newblock {\em Nature}, 530(7588):37--38, 2016.

\bibitem{pincus2016autofluorescence}
Zachary Pincus, Travis~C Mazer, and Frank~J Slack.
\newblock Autofluorescence as a measure of senescence in c. elegans: look to
  red, not blue or green.
\newblock {\em Aging (Albany NY)}, 8(5):889, 2016.

\bibitem{ronneberger2015u}
Olaf Ronneberger, Philipp Fischer, and Thomas Brox.
\newblock U-net: Convolutional networks for biomedical image segmentation.
\newblock In {\em International Conference on Medical image computing and
  computer-assisted intervention}, pages 234--241. Springer, 2015.

\bibitem{sahraeian2019deep}
Sayed Mohammad~Ebrahim Sahraeian, Ruolin Liu, Bayo Lau, Karl Podesta, Marghoob
  Mohiyuddin, and Hugo~YK Lam.
\newblock Deep convolutional neural networks for accurate somatic mutation
  detection.
\newblock {\em Nature communications}, 10(1):1--10, 2019.

\bibitem{stroustrup2013caenorhabditis}
Nicholas Stroustrup, Bryne~E Ulmschneider, Zachary~M Nash, Isaac~F
  L{\'o}pez-Moyado, Javier Apfeld, and Walter Fontana.
\newblock The caenorhabditis elegans lifespan machine.
\newblock {\em Nature methods}, 10(7):665, 2013.

\bibitem{sulston1977post}
John~E Sulston and H~Robert Horvitz.
\newblock Post-embryonic cell lineages of the nematode, caenorhabditis elegans.
\newblock {\em Developmental biology}, 56(1):110--156, 1977.

\bibitem{vaupel1998biodemographic}
James~W Vaupel, James~R Carey, Kaare Christensen, Thomas~E Johnson, Anatoli~I
  Yashin, Niels~V Holm, Ivan~A Iachine, V{\"a}in{\"o} Kannisto, Aziz~A
  Khazaeli, Pablo Liedo, et~al.
\newblock Biodemographic trajectories of longevity.
\newblock {\em Science}, 280(5365):855--860, 1998.

\end{thebibliography}
}

\end{document}